\documentclass[runningheads]{llncs}
\usepackage{graphicx}
\usepackage{subcaption}

\usepackage{tikz}
\usepackage{comment}
\usepackage{amsmath,amssymb} 
\usepackage{color}

\usepackage[accsupp]{axessibility}  

\begin{document}
\pagestyle{headings}
\mainmatter
\def\ECCVSubNumber{049}  

\title{A Geometric-Relational Deep Learning Framework for BIM Object Classification\thanks{This work was supported by the National Key Research and Development Program of China (2021YFB1600303).}} 

\titlerunning{A Geometric-Relational Deep Learning Framework}
%
\author{Hairong Luo\inst{1,2} \and
Ge Gao\inst{1,2} \and
Han Huang\inst{1,2} \and
Ziyi Ke\inst{1} \and
Cheng Peng\inst{1} \and
Ming Gu\inst{1,2}}
\authorrunning{H. Luo et al.}
%
\institute{School of Software, Tsinghua University, Beijing, China \and
Beijing National Research Center for Information Science and Technology(BNRist), Tsinghua University, Beijing, China\\
\email{luohr22@mails.tsinghua.edu.cn, gaoge@tsinghua.edu.cn, h-huang20@mails.tsinghua.edu.cn, ziyike@mail.tsinghua.edu.cn, chengpengeace@gmail.com, guming@mail.tsinghua.edu.cn} 
}
\maketitle

\begin{abstract}
Interoperability issue is a significant problem in Building Information Modeling (BIM). Object type, as a kind of critical semantic information needed in multiple BIM applications like scan-to-BIM and code compliance checking, also suffers when exchanging BIM data or creating models using software of other domains. It can be supplemented using deep learning. Current deep learning methods mainly learn from the shape information of BIM objects for classification, leaving relational information inherent in the BIM context unused. To address this issue, we introduce a two-branch geometric-relational deep learning framework. It boosts previous geometric classification methods with relational information. We also present a BIM object dataset —— IFCNet++, which contains both geometric and relational information about the objects. Experiments show that our framework can be flexibly adapted to different geometric methods. And relational features do act as a bonus to general geometric learning methods, obviously improving their classification performance, thus reducing the manual labor of checking models and improving the practical value of enriched BIM models.
\end{abstract}

\section{Introduction}

Interoperability issues, as an essential problem in the application of BIM, still affect the practical value of BIM models. Proper use of the BIM technique without interoperability problems requires BIM data to be shared and exchanged conveniently and undamaged between software of different professions. Now most BIM software support IFC as a standard data exchange schema, which plays a crucial role in enabling interoperability \cite{koo2019using}. 

However, because IFC contains entity and relationship definitions of various AEC subdomains, the schema defined by IFC is complex and redundant \cite{eastman2010exchange}, which causes unacceptable mismatch and reliability problems \cite{eastman2010exchange,olofsson2008case}. Lack of formal logic rigidness in IFC \cite{eastman2010exchange} also makes mapping of BIM elements to IFC types arbitrary and susceptible to misclassifications \cite{koo2019using}. These phenomena are a cause of interoperability problems and hamper the advance of BIM. Fixing erroneous, misrepresented, contradictory, or missing data that appears during model data exchange remains to be laborious and frustrating \cite{bazjanac2007reduction}. This poses a challenge to the reuse of BIM models in downstream tasks. Semantic enrichment techniques \cite{bazjanac2007reduction,belsky2016semantic,daum2014processing,koo2019using,ma2017building,mazairac2013bimql,olofsson2008case,pauwels2016express,pazlar2008interoperability,qin2014deep,venugopal2012semantics} solve the interoperability problem by exploiting existing numeric, geometric, or relational information in the model to infer new semantic information.

Object classification integrity is a fundamental yet critical requirement that needs to be satisfied using semantic enrichment. Object type provides hints about an object's function, location, size, etc. However, IFC does not ensure correct mapping between BIM objects and their corresponding IFC types \cite{ma2017building}. Missing or incorrect object type usually occurs due to the inconsistent definition of an object’s role in different AEC subdomains. Supplementing the object type information can improve the usability and practical value of BIM models.

Deep learning applications have been explored in various fields in recent years, including BIM object classification. By inputting the objects extracted from an IFC file to a trained deep learning model, the model is able to check the integrity of BIM element to IFC class mappings and identify discrepancies \cite{koo2019using}. They first represent BIM objects as pure geometric representations, such as voxels, meshes, 2D views, or point clouds, then classify the objects using 3D geometric learning models like MVCNN \cite{su2015multi}. This approach neglects the relational information between objects in the BIM context, which might also provide guidance. 

Starting from this intuition, we put forward a geometric-relational deep learning framework that learns the geometric and relational features on different branches and fuses them as a unified object descriptor. Particularly, we propose a relational feature extractor and a feature fusion module in the framework. The two modules serve to extract high-level relational features of BIM objects and fuse them with geometric features extracted by the geometric feature extractor, respectively. This framework can be applied to most existing models and robustly boost their performance, because almost all mainstream geometric deep learning models can serve as the geometric feature extractor. We select MVCNN, DGCNN \cite{wang2019dynamic} and MVViT (a 3D deep learning model adapted from Vision Transformer \cite{dosovitskiy2020image}) as geometric feature extractors in the framework and propose three corresponding models, namely Relational MVCNN (RMVCNN), Relational DGCNN (RDGCNN) and Relational MVViT (RMVViT). Experiments show that with the addition of relational features, the BIM object classification abilities of these models are noticeably improved to varying degrees. This proves the efficacy and flexibility of our framework.

As for the data, there is still a lack of BIM object datasets that contain objects' relations in the BIM context. We propose the IFCNet++ dataset to fill this vacancy. We attach selected representative relational features to each BIM object in the dataset, along with their geometric shapes. We use this dataset in all the experiments for training and testing our models.

To sum up, the contributions of this paper are as follows:
\begin{itemize}

    \item[$\bullet$] Proposing a geometric-relational deep learning framework to utilize both geometric and relational information of BIM objects simultaneously for BIM object classification;
    
    \item[$\bullet$] Putting forward three BIM object classification models based on the geometric-relational framework, and achieving better classification results than their baseline models using additional relational information.
    
    \item[$\bullet$] Proposing IFCNet++, a BIM object dataset containing geometric and relational information for BIM object classification task.
    
    \item[$\bullet$] The efficacy and flexibility of our framework to fully exploit the relational information are demonstrated by comprehensive experiments.
\end{itemize}

\section{Related Work}

\subsection{3D Object Recognition}

3D object recognition is a longstanding problem in computer vision and computer graphics. BIM objects can be seen as 3D objects with semantics, so we can utilize existing object recognition methods to classify them. Early works concentrate on designing 3D local feature descriptors to solve a variety of 3D problems \cite{bennamoun1999feature,bronstein2011shape,gao2014view,guo2013rotational,guo2014accurate,lai2011scalable,lei2014efficient,matei2006rapid,mian2006three,mian2006novel,shang2010real,tombari2013performance}, including 3D object recognition. These handcrafted descriptors are required to be descriptive and robust \cite{guo2016comprehensive}, but they have difficulty in deciding a priori what constitutes meaningful information and what is the result of noise and external factors \cite{koo2021automatic}. In recent years, deep learning based methods attract great attention. Wu {\it et al.} \cite{wu20153d} classify objects represented in voxel form using a 3D convolutional network. Su {\it et al.} \cite{su2015multi} propose an architecture that synthesizes a single compact 3D shape descriptor of an object using image features extracted from the object's multiple views. The synthesis is done by a view-pooling layer. Qi {\it et al.} \cite{qi2017pointnet} design a deep learning model that directly takes the point cloud as input to carry out classification or segmentation tasks. This architecture is invariant to input permutation. Wang {\it et al.} \cite{wang2019dynamic} proposed to dynamically generate graph structures based on the input point cloud, and use EdgeConv modules to perform convolution operations on the graphs. Our geometric-relational framework can be applied to these geometric learning methods to boost their performance on BIM object classification.

\subsection{BIM Object Classification}

BIM object classification is a fundamental task of BIM semantic enrichment. Relative methods can be divided into deductive methods and inductive methods \cite{bloch2018comparing}. Deductive methods require the design of explicit and unique rules for the particular object. They usually ensure certainty in conclusions, but the labor of devising rules for all possible pairs of types makes these methods practically intractable \cite{koo2019using}. As an example, Ma {\it et al.} \cite{ma2017building} propose a procedure for establishing a knowledge base that associates objects with their features and relationships, and a matching algorithm based on a similarity measurement between the knowledge base and facts. Machine learning approaches are representative of inductive methods. These methods generate their own rules inductively by optimizing the weights of features in a model \cite{koo2019using}. Koo {\it et al.} \cite{koo2019using} use support vector machines to check the semantic integrity of mappings between BIM elements and IFC classes. 
Kim {\it et al.} \cite{kim2019recognizing} use 2D CNN to classify furniture entities according to their images. 
Koo {\it et al.} \cite{koo2021automatic} compare the classification results of MVCNN and PointNet on wall subtypes and door subtypes. 
Koo {\it et al.} \cite{koo2021geometric} also use the same two models to classify BIM objects in road infrastructure. 
Collins {\it et al.} \cite{collins2021assessing} encode BIM objects using two kinds of graph encodings and utilize a graph convolutional network to create meaningful local features for subsequent classification. Emunds {\it et al.} \cite{emunds2022sparse} propose an efficient neural network based on sparse convolutions to learn from point cloud representation of BIM objects. This study is also dedicated to solving the BIM object classification problem using deep learning methods.

\subsection{BIM Object Datasets}
Sufficiently large and comprehensive datasets are the key to the training of deep learning models. Currently, there are two relevant datasets, IFCNet \cite{emunds2021IFCnet} and BIMGEOM \cite{collins2021assessing}. IFCNet is a dataset of single-entity IFC files spanning a broad range of IFC classes containing both geometric and single-object semantic information. BIMGEOM is assembled of building models from both industry and academia. It consists of structural elements, equipment and interior furniture types. Both datasets don't contain any relational information of objects, thus not suitable for our framework. We propose IFCNet++ which involves certain relationships between objects and use it to train our models.

\section{Geometric-Relational Deep Learning Framework}

\begin{figure}
\centering
\includegraphics[height=5.5cm]{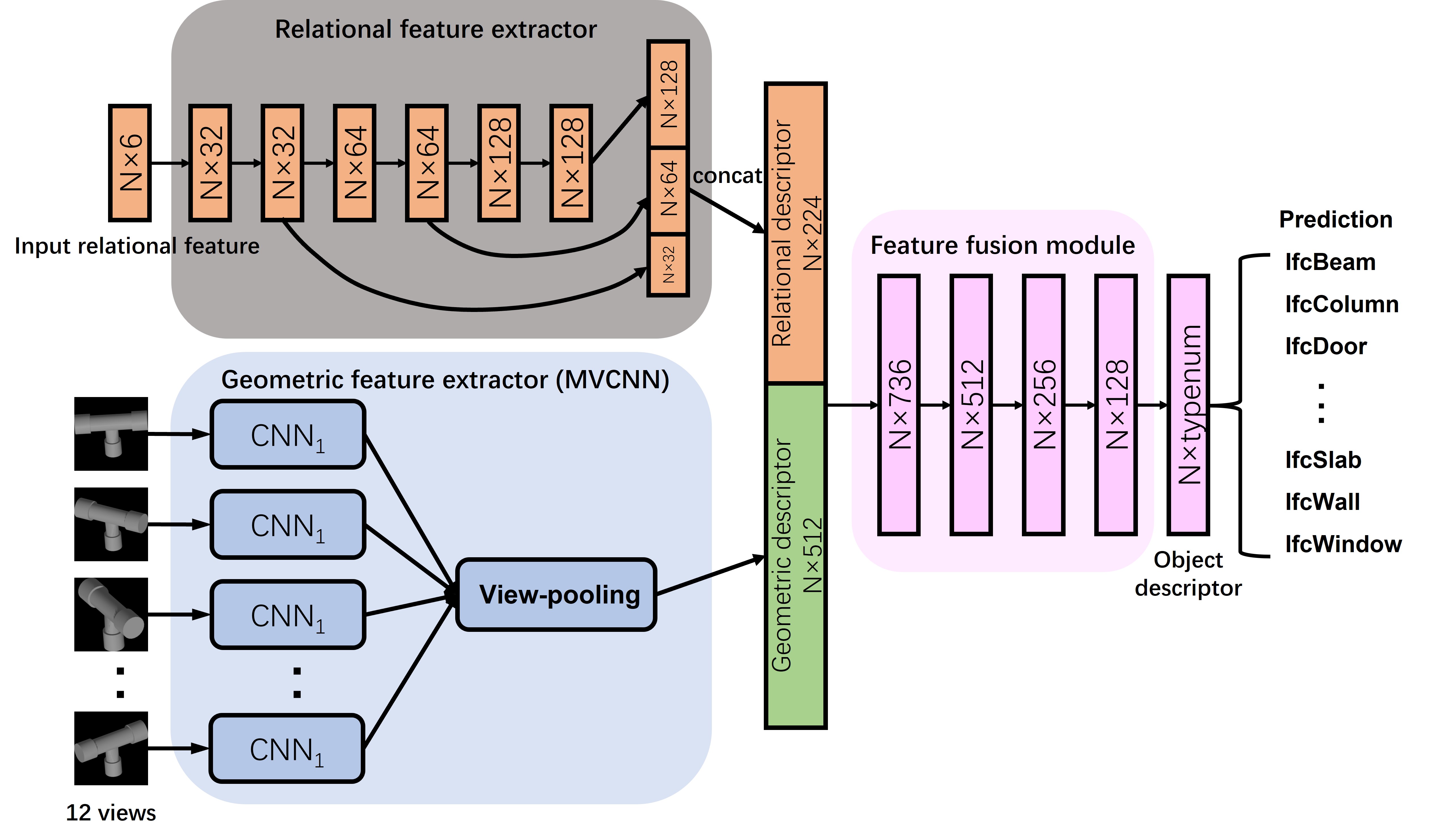}
\caption{Geometric-relational framework overview. The geometric feature extractor represents a common geometric learning backbone. In this figure we use the MVCNN backbone. It extracts the geometric feature of the BIM object from its shape representation. The relational feature extractor extracts relational features from raw relational data. It also connects low-level features to high-level features to form the complete relational feature. The feature fusion module fuses these two kinds of features and outputs the object descriptor, which can be sent to a classifier to get the corresponding type label of the object}
\label{fig:framework}
\end{figure}

The target of our research is to create a reasonable deep learning framework to simultaneously learn the geometric and relational information for BIM object classification. Object shape data and relational data are usually presented in different forms, which makes learning these features using a single network branch difficult. So we adopt a two-branch method intuitively. The two branches can use different network architectures to process the two kinds of information and extract a geometric descriptor and a relational descriptor of the object. These descriptors are fused to get a unified object descriptor.

We show the overview of our geometric-relational deep learning framework in Fig.~\ref{fig:framework}. It consists of three main modules, i.e. geometric feature extractor, relational feature extractor and feature fusion module. The geometric feature extractor learns to represent a BIM object's shape as a geometric descriptor. The relational feature extractor extracts relational features of different levels and connects them to form the relational descriptor. The feature fusion module mixes the two descriptors and outputs the final object descriptor. This descriptor is a more complete abstraction of the BIM object than a pure geometric descriptor, and can be used by a classifier to perform classification more accurately. Next, we will introduce the module designs and implementation details of our framework.

\subsection{Module Designs}

In the following, we make a detailed explanation of the design intuitions and detailed structures of the three modules in our framework.

{\bf Geometric feature extractor} serve to learn from objects' shapes. It takes the raw geometric data of BIM objects as input and outputs their high-level geometric descriptors. A BIM object's shape can be represented in various forms, like multiple views, point clouds, voxels, etc. To properly extract geometric features from these representations, we do not fix the geometric feature extractor to a certain design. Instead, any geometric deep learning method that can extract a geometric descriptor from an object can be the geometric extractor. This provides flexibility to the geometric input form and our framework design. It also empowers our framework to boost geometric methods' of any kind with relational information. The classification result of our framework would also be improved as better geometric deep learning models are proposed. In our implementation, we use MVCNN and MVViT to learn from objects' multi-view representation and DGCNN to learn from the point cloud representation.

{\bf Relational feature extractor} is used to learn the relation pattern of each BIM object type. It takes as input a 1D vector to represent an object's interaction with the context and outputs the relational descriptor. We use an MLP with batch normalization and ReLU to gradually extract high-level relational features. In the meantime, we have observed that the original input vectors also show some simple distribution patterns that may be helpful to represent the object's type. This indicates that low-level relational features may also be instructive to the classification process. Therefore we directly connect features of former layers to the feature of the last layer to form the final relational descriptor.

{\bf Feature fusion module} merges the above two branches and further studies a unified BIM object descriptor using the extracted geometric feature and relational feature. It is designed as an MLP with batch normalization and ReLU, too, but without connections between different layers. It first concatenates the two descriptors, then uses the MLP to extract the object descriptor, which can be sent to a classifier.

\subsection{Implementation Details}

According to the observation that the geometric aspect of a BIM object contains more useful information and can provide more clues about the object's type label than its relational aspect, we design the geometric branch as the main branch and the relational branch as a supplement. Specifically, the geometric feature extractor is more complicated than the relational counterpart. And the geometric descriptor size is larger than the relational descriptor size. This viewpoint can be observed in the following implementation details.

Because we directly utilize existing geometric learning methods as the geometric feature extractor, we skip this part and begin the introduction with the relational feature extractor. As shown in Fig.~\ref{fig:framework}, this branch consists of 6 linear layers. It gradually embeds the input 6-dimensional relation vector into a high-level feature space of 128 dimensions. We also concatenate the feature of the second and the fourth layer to the feature of the last layer to form the final relational descriptor. The total length of the descriptor is 224. The simple design of the relational branch guarantees that it takes up less computational resources, verifying the viewpoint in the previous paragraph. 

The feature fusion module is composed of four linear layers. It maps the concatenation of the geometric descriptor and the relational descriptor down to a 128-dimension feature space. The feature length of the first module layer is the same as the length of the input, so it may vary as the geometric feature extractor changes. The second to the last layers output features of fixed length. Detailed length information is shown in Fig.~\ref{fig:framework}. We use a linear layer with a softmax operation as the classifier, which outputs the predicted probability of each BIM object type.

\section{Implementation}

\subsection{Relational Models}
We select MVCNN and DGCNN as our baseline for multi-view based method and point cloud based method, respectively. To test our framework on a Transformer based model to explore its generalization, we modify Vision Transformer to a multi-view based geometric learning method. We use its backbone to extract a classification token for each view, and max-pool the tokens to obtain the geometric feature descriptor of the object. The model is called Multi-view Vision Transformer or MVViT. We transform these three models into corresponding relational models using our framework, which are called RMVCNN, RDGCNN and RMVViT, respectively. 

\subsection{Training Configurations}
All the models are implemented based on PyTorch and trained using NVIDIA RTX 3090. The batch size is 64. Adam optimizer is used with $\beta_1$=0.9, $\beta_2$=0.999 and $\epsilon$=1e-8. We use cross-entropy loss as the loss function.

In the implementation, we adopt pre-trained ResNet34 \cite{he2016deep} as the backbone network of MVCNN and RMVCNN. For MVViT and RMVViT, we use pre-trained ViT-Base as the backbone. Each baseline and its relational model share the same learning rate and weight decay. The training epochs are tailored for each model to get relatively good performance.

\section{IFCNet++ Dataset}

\subsection{Dataset Overview}

\begin{figure}
\centering
\includegraphics[height=3.0cm]{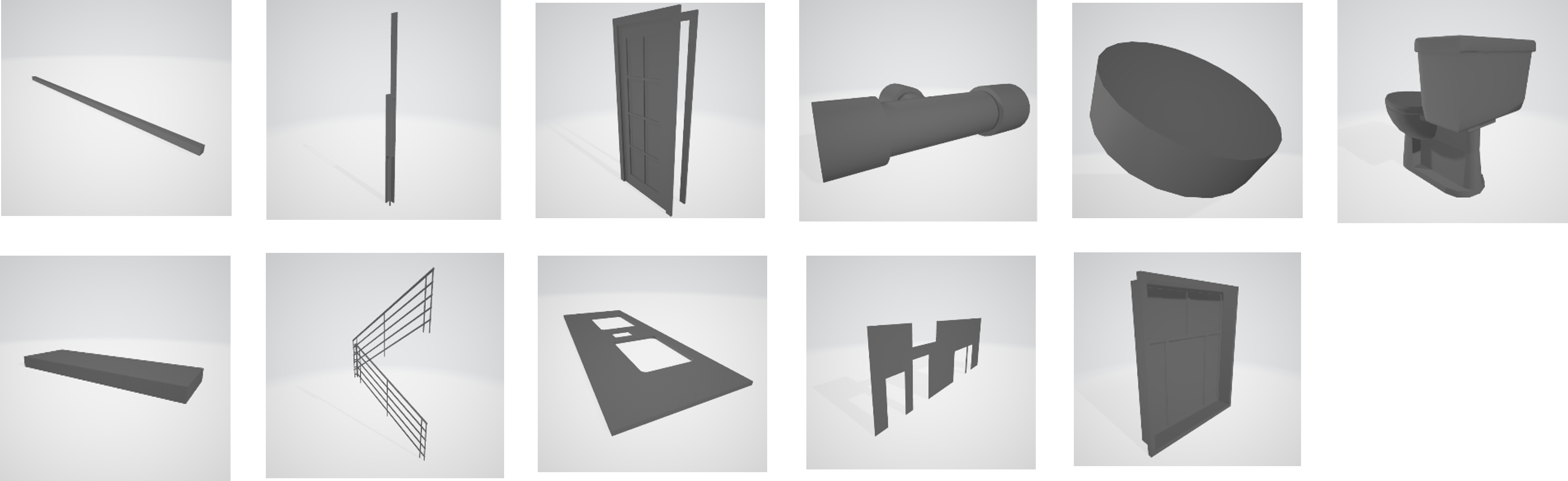}
\caption{Dataset samples: (from left to right in reading order) IfcBeam, IfcColumn, IfcDoor, IfcFlowFitting, IfcFlowSegment, IfcFlowTerminal, IfcPlate, IfcRailing, IfcSlab, IfcWall, IfcWindow}
\label{fig:typeexamples}
\end{figure}

We focus on BIM object classification based on supervised learning. Supervised learning methods use labeled datasets to train models to learn data distribution. There are two representative BIM object datasets, which are IFCNet and BIMGEOM. These two datasets present BIM objects using file formats such as ifc, obj and png. Except for the ifc format, which contains varying degrees of single-object semantics like size and material, other formats only contain geometric information about the objects. However, BIM models are rich in semantics. That means BIM objects are not pure geometric shapes. They exist in the building context and have various spatial and topological relations with other objects. So these relationships should be recorded to preserve the BIM nature when making BIM object datasets. Therefore, we propose IFCNet++ as an enhanced BIM object dataset. It contains BIM objects' geometric shapes and interactions with other objects. We collect 9228 objects belonging to 11 most common types in the dataset. The 11 types are selected based on their appearance frequency and importance in BIM models. They can cover most objects of interest in a BIM model, and cover object types in both architectural models and MEP models. An overview of the dataset is shown in Fig.~\ref{fig:typeexamples} and Table~\ref{table:datasetoverview}.

\setlength{\tabcolsep}{4pt}
\begin{table}
\begin{center}
\caption{Data distribution overview of IFCNet++ dataset}
\label{table:datasetoverview}
\begin{tabular}{lccc}
\hline\noalign{\smallskip}
BIM object type & Training & Testing & Total\\
\noalign{\smallskip}
\hline
\noalign{\smallskip}
IfcBeam & 66 & 27 & 93\\
IfcColumn & 64 & 27 & 91\\
IfcDoor & 939 & 402 & 1341\\
IfcFlowFitting & 103 & 43 & 146\\
IfcFlowSegment & 115 & 49 & 164\\
IfcFlowTerminal & 273 & 116 & 389\\
IfcPlate & 1400 & 600 & 2000\\
IfcRailing & 210 & 90 & 300\\
IfcSlab & 929 & 397 & 1326\\
IfcWall & 1400 & 600 & 2000\\
IfcWindow & 965 & 413 & 1378\\
\noalign{\smallskip}
\hline
\noalign{\smallskip}
Total & 6464 & 2764 & 9228\\
\hline
\end{tabular}
\end{center}
\end{table}
\setlength{\tabcolsep}{1.4pt}

\subsection{Relational Feature Design}

IFCNet++ focuses on recording relational information of objects. We extract relational information from relational items in IFC files. IFC schema has defined abundant relationships among objects. But in most cases, few relationships have been implemented in an IFC file. Some relationships may also get lost because of interoperability problems. Therefore, we select four subtypes of IfcRelationship that appear in most IFC files and are easy to extract, i.e. IfcRelConnectsElement, IfcRelFillsElement, IfcRelAggregates and IfcRelVoidsElement.

Another problem is how to represent these relationships in the dataset. A direct idea is to build a connected relation graph of a BIM model. That is, each object represents a node and each relational item represents a connecting edge. However, the selected relationships usually exist in local regions. They can not connect all the objects in a whole graph, so we discard the graph based solution. Instead, we adopt a counting method and attach a simple 1D vector to each BIM object. Each vector consists of six numbers. Each number represents how many times an object is quoted in a certain relationship attribute. The six numbers correspond to the following six attributes:

\begin{itemize}
    \item[$\bullet$] IfcRelConnectsElement.RelatingElement
    \item[$\bullet$] IfcRelConnectsElement.RelatedElement
    \item[$\bullet$] IfcRelAggregates.RelatingObject
    \item[$\bullet$] IfcRelAggregates.RelatedObjects
    \item[$\bullet$] IfcRelVoidsElement.RelatingBuildingElement
    \item[$\bullet$] IfcRelFillsElement.RelatedBuildingElement
\end{itemize}

Despite being simple, this vectorized form has a strong representative ability for the relational information, as the simple local relational structure can be well represented by this counting approach. Besides, this form is convenient for later processes using MLP. It also maintains a good trade-off between data size and the ability to represent relational information.   

\subsection{Data Collection and Processing}

We collected the BIM objects in IFCNet++ from more than ten IFC files. We first split the model into individual objects. Then we extract all the objects of interest in obj format to get their geometric representation. Deduplication is performed on the collected objects. We consider an object as a duplication if it can overlap with another object after translation and rotation. Next, we extract the four selected relationship items and count how many times an object is quoted by a certain relationship attribute. Finally, we attach the counted vectors to the corresponding objects. The collected object distribution is very unbalanced across object types. For example, wall objects tend to appear in large amounts for a BIM model. So we randomly select at most 2000 objects for a certain object type. We split the training set and the testing set by a ratio of 7:3. 

\begin{figure}
    \centering
    \begin{subfigure}[b]{0.65\textwidth}
        \centering
        \includegraphics[width=\textwidth]{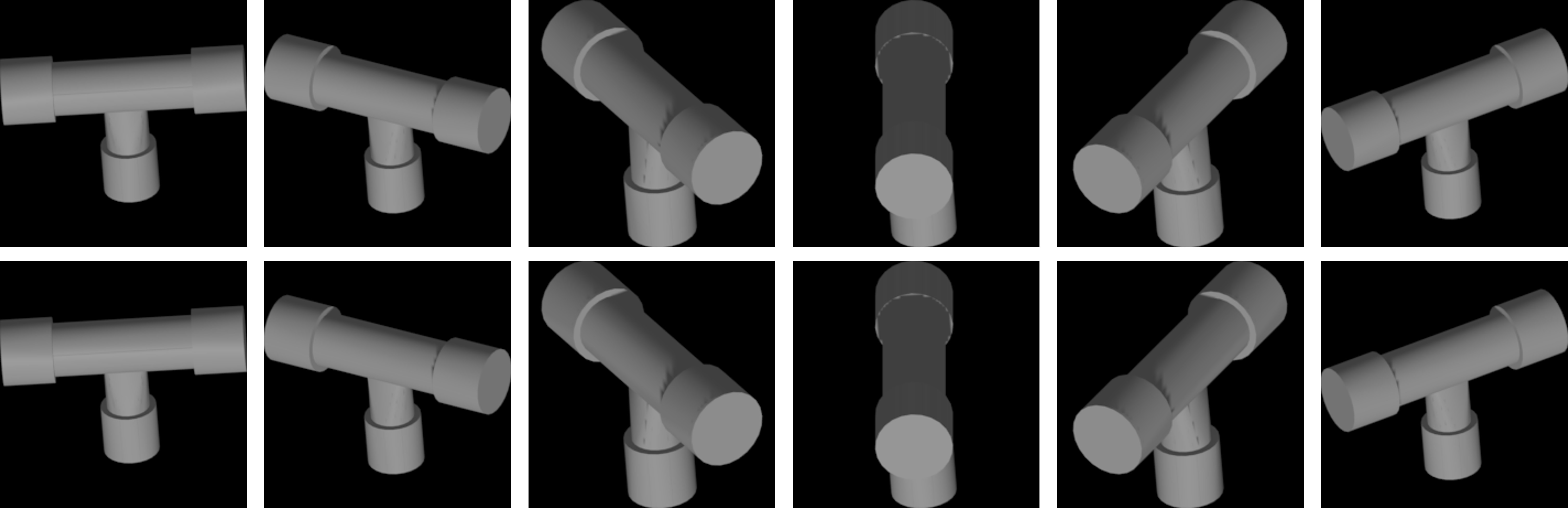}
        \caption{12-view representation}
        \label{fig:multiview}
    \end{subfigure}
    ~ 
    \begin{subfigure}[b]{0.29\textwidth}
        \centering
        \includegraphics[width=\textwidth]{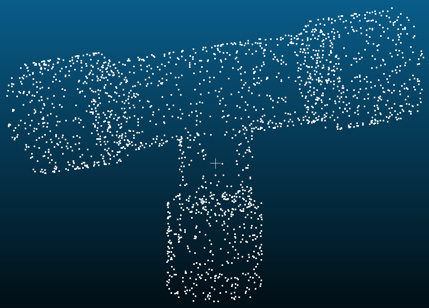}
        \caption{Point cloud}
        \label{fig:pointcloud}
    \end{subfigure}
    \caption{Two representations of an IfcFlowFitting object in IFCNet++}
    \label{fig:representations}
\end{figure}

Besides, we need to further process the obj files to get the proper input format of multi-view based and point cloud based methods. We render each object to get a 12-view representation using the rendering method of \cite{su2018deeper}. We also convert objects to point cloud form using the code in \cite{emunds2021IFCnet}. We show an example of our 12-view representation and point cloud representation in Fig.~\ref{fig:representations}.

\section{Experiments}
In this section, we show the experimental results of our framework. Experiments on the three pairs of baseline and relational models show the performance boost gained by taking relations into the learning process.

\subsection{Testing Metrics}

We test the models on the test set of IFCNet++ and show their classification results in Table~\ref{table:testmetrics}.

\setlength{\tabcolsep}{4pt}
\begin{table}
\begin{center}
\caption{Classification results of our trained models on the IFCNet++ test set}
\label{table:testmetrics}
\begin{tabular}{lccccc}
\hline\noalign{\smallskip}
Model & Accuracy & Balanced accuracy & Precision & Recall & F1 score\\
\noalign{\smallskip}
\hline
\noalign{\smallskip}
MVCNN & 0.9732 & 0.9549 & 0.9742 & 0.9732 & 0.9734\\ 
DGCNN & 0.9801 & 0.9536 & 0.9809 & 0.9801 & 0.9802\\ 
MVViT & 0.9797 & 0.9527 & 0.9812 & 0.9797 & 0.9799\\
\noalign{\smallskip}
\hline
\noalign{\smallskip}
RMVCNN & {\bf 0.9917} & {\bf 0.9750} & {\bf 0.9918} & {\bf 0.9917} & {\bf 0.9916}\\
RDGCNN & 0.9902 & 0.9624 & 0.9906 & 0.9902 & 0.9903\\
RMVViT & 0.9841 & 0.9581 & 0.9857 & 0.9841 & 0.9842\\
\hline
\end{tabular}
\end{center}
\end{table}
\setlength{\tabcolsep}{1.4pt}

The baselines can already reach a high precision of 97\% and a balanced accuracy of 95\% merely utilizing the geometric information. This illustrates that most BIM objects can be correctly classified by their shapes. However, with the addition of relational features, the three relational models can get better results on all the metrics than their corresponding baselines. This intuitively shows that the composition of selected relationships can effectively represent the local relation patterns of each object type. And our framework can learn these patterns and fuse them with geometric features to refine the object descriptors.

Noticeably, even though MVCNN doesn't perform very well relative to the other two baselines, RMVCNN not only performs best on all the metrics, but also gains the most improvement with each metric improved by about 2\%. RDGCNN and RMVViT have been improved by about 1\% and 0.5\% on each metric, respectively. This shows even with the same input relational features, certain geometric models can gain better improvement using our framework. The relational feature space fuses best with the geometric feature space learned by MVCNN to gain the most improvement. So the key to better classification results is to find a proper geometric method that fits our framework well.

\setlength{\tabcolsep}{4pt}
\begin{table}
\begin{center}
\caption{Classification accuracy of RMVCNN organized by object types}
\label{table:testaccuracy}
\begin{tabular}{lccc}
\hline\noalign{\smallskip}
Object type & Total & Correctly classified & Accuracy(\%)\\
\noalign{\smallskip}
\hline
\noalign{\smallskip}
IfcBeam & 27 & 27 & 100.0\\ 
IfcColumn & 27 & 24 & 88.9\\
IfcDoor & 402 & 399 & 99.3\\
IfcFlowFitting & 43 & 43 & 100.0\\
IfcFlowSegment & 49 & 43 & 87.8\\
IfcFlowTerminal & 116 & 116 & 100.0\\
IfcPlate & 600 & 598 & 99.7\\
IfcRailing & 90 & 89 & 98.9\\
IfcSlab & 397 & 390 & 98.2\\
IfcWall & 600 & 599 & 99.8\\
IfcWindow & 413 & 413 & 100.0\\
\noalign{\smallskip}
\hline
\noalign{\smallskip}
Total & 2764 & 2741 & 99.2\\
\hline
\end{tabular}
\end{center}
\end{table}
\setlength{\tabcolsep}{1.4pt}

We list the classification accuracy of RMVCNN organized by object types in Table~\ref{table:testaccuracy} to further explore its classification ability. RMVCNN performs well on 9 of the 11 types, reaching an accuracy higher than 98\%. This means RMVCNN can effectively learn geometric and relational features of most types. Accuracy on the other two types is slightly lower than 90\%. This may be partially explained by the small quantities of training samples of these types.

\subsection{Confusion Rate}

The geometric learning baselines are prone to be confused by object types that contain geometrically similar objects. To quantitatively analyze this trend, we propose the notion of confusion rate. Suppose A and B are two object types, we define a model's confusion rate between a pair of types A and B as:
\begin{equation}
    c = \frac{m_{AB} + m_{BA}}{n_A + n_B} .
\end{equation}
Here $m_{AB}$ is the amount of BIM objects of type A misclassified as type B. $m_{BA}$ is the amount of BIM objects of type B misclassified as type A. $n_A$ and $n_B$ represent the amounts of objects of type A and type B in the test set.

We compute the confusion rates of MVCNN and RMVCNN on the test set. We then sort the type pairs in descending order of the confusion rates of MVCNN and list the results of the first ten pairs in Table~\ref{table:confusionrate}.

\setlength{\tabcolsep}{4pt}
\begin{table}
\begin{center}
\caption{Confusion rates of MVCNN and RMVCNN. The first ten type pairs sorted in descending order of MVCNN's confusion rates are selected}
\label{table:confusionrate}
\begin{tabular}{llcc}
\hline\noalign{\smallskip}
Type 1 & Type 2 & MVCNN(\%) & RMVCNN(\%)\\
\noalign{\smallskip}
\hline
\noalign{\smallskip}
IfcColumn & IfcFlowSegment & 9.2 & 3.9\\
IfcBeam & IfcFlowSegment & 5.3 & 5.3\\
IfcDoor & IfcWall & 1.2 & 0\\
IfcPlate & IfcSlab & 1.1 & 0.2\\
IfcSlab & IfcWall & 0.9 & 0.4\\
IfcBeam & IfcWall & 0.6 & 0\\
IfcFlowFitting & IfcFlowTerminal & 0.6 & 0\\
IfcFlowSegment & IfcFlowTerminal & 0.6 & 0\\
IfcFlowTerminal & IfcPlate & 0.6 & 0\\
IfcColumn & IfcWall & 0.5 & 0.3\\
\hline
\end{tabular}
\end{center}
\end{table}
\setlength{\tabcolsep}{1.4pt}

In the ten type pairs that MVCNN is prone to confuse, RMVCNN's confusion rates have obviously reduced on nine of them. The confusion rates of five type pairs have even been reduced to zero. It shows that for most type pairs that share similar object shapes, relational information can help the model find the difference between them according to the union of their shapes and context information. What's more, the highest confusion rate has come down from 9.2\% of MVCNN to 5.3\% of RMVCNN. This implies that relational information plays an important part in lowering both the average level and the upper limit of confusion rates. Our framework performs well in alleviating the model's confusion situations.

\subsection{Corrected Classification Results}
\begin{figure}
    \centering
    \begin{subfigure}[t]{0.3\textwidth}
        \centering
        \includegraphics[width=\textwidth]{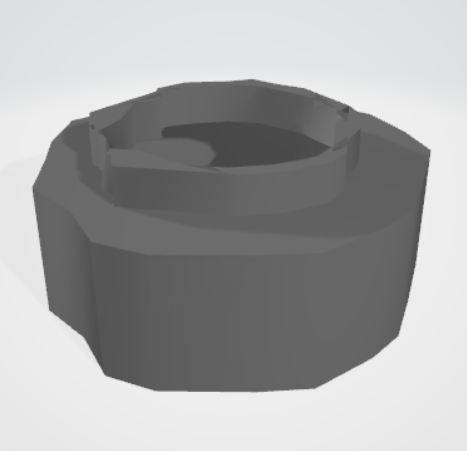}
        \caption{IfcFlowTerminal misclassified as IfcFlowFitting}
        \label{fig:example_IFCflowterminal}
    \end{subfigure}
    ~ 
    \begin{subfigure}[t]{0.315\textwidth}
        \centering
        \includegraphics[width=\textwidth]{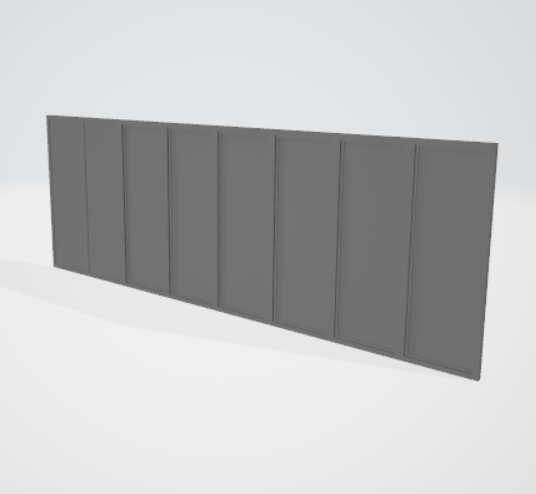}
        \caption{IfcWindow misclassified as IfcRailing}
        \label{fig:example_IFCwindow}
    \end{subfigure}
    ~ 
    \begin{subfigure}[t]{0.3\textwidth}
        \centering
        \includegraphics[width=\textwidth]{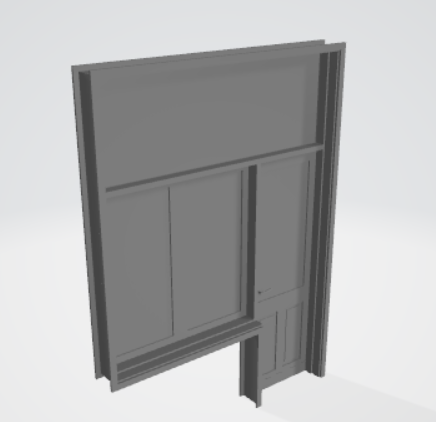}
        \caption{IfcDoor misclassified as IfcWindow}
        \label{fig:example_IFCdoor}
    \end{subfigure}
    \caption{Demonstration of three BIM objects misclassified by MVCNN but correctly classified by RMVCNN}
    \label{fig:examples}
\end{figure}

We display some of the BIM objects that are misclassified by MVCNN but correctly classified by RMVCNN in Fig.~\ref{fig:examples}. Fig.~\ref{fig:example_IFCflowterminal} shows an IfcFlowTerminal. Because its shape looks like a joint of two pipelines, it is misclassified as an IfcFlowFitting by MVCNN. However, with the assistance of its relational features, RMVCNN can judge its type correctly. A similar problem happens when MVCNN tries to classify the IfcWindow in Fig.~\ref{fig:example_IFCwindow}. According to the thin columns on its surface, this window looks much like a straight railing, so MVCNN classifies it as an IfcRailing wrongly. This mistake is also avoided by RMVCNN. The IfcDoor in Fig.~\ref{fig:example_IFCdoor} is misclassified as IfcWindow by MVCNN for its window-like frame and also gets corrected by RMVCNN. These examples clearly demonstrate our framework's advantage in distinguishing BIM objects with misleading geometric shapes.

\subsection{Computational Cost}
To figure out the trade-off between performance boost and computational cost introduced by our framework, we list the number of parameters and calculations of MVCNN and RMVCNN in Table~\ref{table:computationalcost}.

\setlength{\tabcolsep}{4pt}
\begin{table}
\begin{center}
\caption{Number of parameters and calculations of MVCNN and RMVCNN}
\label{table:computationalcost}
\begin{tabular}{lcc}
\hline\noalign{\smallskip}
Model & Params(M) & MACs(M)\\
\noalign{\smallskip}
\hline
\noalign{\smallskip}
MVCNN & 21.290 & 44049.020\\
RMVCNN & 22.407 & 44050.135\\
\noalign{\smallskip}
\hline
\noalign{\smallskip}
Cost introduced & 1.117 & 1.115\\
\hline
\end{tabular}
\end{center}
\end{table}
\setlength{\tabcolsep}{1.4pt}

By applying our framework, we only introduce about 1.1~M params (5\%) and 1.1~M MACs (0.0025\%) to RMVCNN relative to MVCNN. Considering the fact that MVCNN can already reach a very high classification accuracy, further improvement is hard to achieve. However, our framework uses a relatively low price to push its performance boundary by an obvious margin. It avoids adding a huge amount of computational cost like some large-scale models. This result shows the superiority and efficiency of our method.

\subsection{Ablation Study}
We conduct ablation studies on RMVCNN and remove each of the three component modules to look into their contributions. When removing the geometric or relational feature extractor, we also abandon the corresponding input information. The results are shown in Table~\ref{table:ablation}. When removing the geometric feature extractor, the fundamental part of the framework, the results are very poor and could not be trusted. Because the relational feature extractor acts as an auxiliary part, the results of the framework without it look good but slightly suffer. The experiment without the feature fusion module shows better results, but could not reach the results of the full RMVCNN. This illustrates that the fusion process does help in extracting a more compact and representative object descriptor.

\setlength{\tabcolsep}{4pt}
\begin{table}
\begin{center}
\caption{Ablation studies. We remove each of the three modules in RMVCNN to validate their contribution in recognizing BIM objects}
\label{table:ablation}
\begin{tabular}{lccccc}
\hline\noalign{\smallskip}
Model & Accuracy & Balanced accuracy & Precision & Recall & F1 score\\
\noalign{\smallskip}
\hline
\noalign{\smallskip}
w/o geometric & 0.5380 & 0.2839 & 0.4748 & 0.5380 & 0.4030\\
w/o relational & 0.9805 & 0.9667 & 0.9813 & 0.9805 & 0.9807\\
w/o fusion & 0.9881 & 0.9695 & 0.9883 & 0.9881 & 0.9881\\
full RMVCNN & 0.9917 & 0.9750 & 0.9918 & 0.9917 & 0.9916\\
\hline
\end{tabular}
\end{center}
\end{table}
\setlength{\tabcolsep}{1.4pt}

\section{Conclusion}

We focus on the BIM object classification problem to ease the interoperability problem of BIM software. We first propose a two-branch geometric-relational deep learning framework. It introduces relational information of BIM objects to assist pure geometric deep learning methods which neglect the relational information inherent in BIM models. Geometric descriptors and relational descriptors are extracted by the two branches, respectively. They are mixed by the feature fusion module to generate the final object descriptors. Our design of the geometric feature extractor makes the framework applicable to most existing geometric learning methods, including CNN based and Transformer based methods. And the framework can always boost their classification performance to different degrees. It shows the efficacy and flexibility of our framework. 

Then, to fill the vacancy in BIM object datasets with relationships, we collect the IFCNet++ dataset. It contains BIM objects' geometric representation and certain local relationships. The relational information is stored in a vectorized form, easy to be processed. Though simple, the relationships are representative enough to help the geometric methods achieve a performance gain. 

We follow our framework to put forward three relational models based on different geometric learning baselines and carry out experiments on them. We found that only with little additional cost introduced, our relational models can utilize objects' relations to better distinguish between BIM types with similar looks. They compensate for the weakness of pure geometric-based methods. 

The limitation of our research lies in three aspects:

\begin{itemize}
    \item[$\bullet$] The relationships of interest are explicitly presented in IFC files. They may also be wrongly labeled or lost, affecting the performance of our method.
    \item[$\bullet$] IFCNet++ dataset only covers major BIM object types. It still needs to be enriched with more object types, so that it can be used to train a more generalized deep learning model.
    \item[$\bullet$] The BIM object types are coarsely defined. They do not include subtype information of objects to provide fine-grained type information and domain-specific knowledge required in some AEC subdomains.
\end{itemize}



%
%
\bibliographystyle{splncs04}
\bibliography{output}

\end{document}